# Image anomaly detection and prediction scheme based on SSA optimized ResNet50-BiGRU model

Qianhui Wan[1], Zecheng Zhang[2], Liheng Jiang[3], Zhaoqi Wang[4], Yan Zhou[5*]

[1]Department of Mathematics, University of California Davis, Davis, California, 95616, United States;qhswan@ucdavis.edu

[2]New York University, Brooklyn, New York, 11201, United States; roderickzzc@gmail.com

[3]New York University, Manhattan, New York, 10012, United States; lj1070@nyu.edu

[4]Viterbi School of Engineering, University of Southern California, Los Angeles, California, 90089, United States; josiewangwzq@gmail.com

[5*]Northeastern University San Jose, San Jose, California, 95131, United States; yanzhouca23@gmail.com

*Corresponding Author: yanzhouca23@gmail.com

## ABSTRACT

Image anomaly detection is a popular research direction, with many methods emerging in recent years due to rapid advancements in computing. The use of artificial intelligence for image anomaly detection has been widely studied. By analyzing images of athlete posture and movement, it is possible to predict injury status and suggest necessary adjustments. Most existing methods rely on convolutional networks to extract information from irrelevant pixel data, limiting model accuracy. This paper introduces a network combining Residual Network (ResNet) and Bidirectional Gated Recurrent Unit (BiGRU), which can predict potential injury types and provide early warnings by analyzing changes in muscle and bone poses from video images. To address the high complexity of this network, the Sparrow search algorithm was used for optimization. Experiments conducted on four datasets demonstrated that our model has the smallest error in image anomaly detection compared to other models, showing strong adaptability. This provides a new approach for anomaly detection and predictive analysis in images, contributing to the sustainable development of human health and performance.

## 1. Introduction

Abnormality detection and analysis of medical images for muscles and bones has been widely studied, and people often cause muscle or bone damage due to incorrect pos-ture during exercise. As recent injury statistics and research show [1], certain body parts are particularly vulnerable to severe damage, such as musculoskeletal injuries, particu-larly in the areas of ankles and ligaments. To address the sport injuries caused by incor-rect postures, many methods have been proposed. For example, Convolutional Neural Network (CNN) [2] based methods have been widely applied in the field of sports science. CNNs excel at dealing with spatial distribution by learning to extract spatial features through convolutional kernels. Their strength lies in handling high-dimensional spatial data, making them particularly suited for images and other types of data.

The combination of the ResNet [3] method and LSTM [4] models has been applied in long sequence prediction. These models perform well when dealing with sequence or spatial data. For instance, PoseNet [5], OpenPose [6]. These models learn from a large amount of image or video data with posture annotations, ena-bling them to identify and assess the correctness of human posture in images or videos, assisting athletes in timely correction of incorrect postures, and thereby preventing sports injuries. However, these methods still face issues such as overfitting and loss of infor-mation when training with a large amount of sports video data, and their detection per-formance on athletes' postures is not satisfactory.

In response to these issues, we propose a method based on the improved Res-Net50-BiGRU model. This method integrates the Residual Connection Network (ResNet), time series model (BiGRU) [7], and Sparrow Search Algorithm (SSA) [8] for optimization. It effectively predicts the changes of muscles and bones during exercise, formulates injury prevention measures, and uses the SSA algorithm to optimize data, generate data specific to specific situations, and provide strong reference and sup-port for decision-making.

The contribution points of this paper are as follows:

• Our proposed method shows excellent performance on multiple video datasets of muscular and skeletal changes, accurately predicting the state of muscular and skel-etal changes in people during sports activities, which can help formulate appropriate sports injury prevention measures.

• The ResNet50-BiGRU model proposed in this study showed excellent efficacy in es-timating muscle and bone changes, and achieved significant improvements in model optimization using the Sparrow Search Algorithm (SSA).

• By comparing the performance of a range of different models in the domain of anom-aly detection in medical images, our experiments show that the SSA-optimized Res-Net50-BiGRU model produces better results than other model architectures.

## 2. Related Work

### 2.1 Convolutional Neural Networks

Convolutional Neural Networks (CNNs) [9], an algorithm from the realm of deep learning, have demonstrated exceptional performance in handling image data in the form of pixel grids. CNNs leverage convolution computations, a mathematical operation that cross-multiplies and sums an input image with a weight matrix known as a convolution kernel or filter, thereby generating new feature maps.

CNNs are frequently deployed for posture estimation tasks, with the central premise being the identification of keypoints (e.g., joints or specified body parts) within images and inferring their spatial arrangements[10]. More specifically, CNNs are capable of under-standing the complex correspondence between keypoints and image pixels, thereby achieving high precision in keypoint detection. In the continuous tracking of athletes' postures, it is possible to identify patterns of behavior leading to sports injuries. For in-stance, an athlete running with an ankle angle frequently deviating from the normal range might be at heightened risk of injury. By learning the video motion characteristics of ath-letes, CNNs can aid in predicting and preventing sports injuries.

However, traditional CNNs lack temporal continuity processing capability and thus are unable to handle information in a temporal sequence. Consequently, when processing continuous video frames, CNNs overlook the temporal relationship between frames. Fur-thermore, video data usually comprises a multitude of frame images, the processing of which necessitates considerable computational resources and storage space[11]. Therefore, handling athletes' motion video data with CNNs can result in computational overhead and insufficient storage issues. To overcome these challenges, researchers have developed some specialized models for video data, such as 3D Convolutional Neural Networks (3D-CNNs) and Temporal Convolutional Networks (TCNs), as well as hybrid models like CNN-LSTM or ResNet50-LSTM that combine convolutional networks (for handling spa-tial information) and recurrent networks (for managing temporal information). These models aim to consider both spatial and temporal information when handling video data.

## 2.2 LSTM Model

Long Short-Term Memory (LSTM) is a specific type of Recurrent Neural Network (RNN)[12] architecture that was designed to tackle the problem of long-term dependencies in sequence prediction tasks. Traditional RNNs have difficulty learning to connect infor-mation that is many time steps apart due to the vanishing gradient problem. LSTM net-works, on the other hand, possess a sophisticated gating mechanism that selectively lets information through to mitigate this issue. This makes LSTMs more capable of handling sequences of data over longer periods, such as those often found in videos of sports per-formances.

In the domain of sports athlete posture analysis and injury detection, LSTM models can play a crucial role[13]. Once the spatial features have been extracted from the video data using networks such as ResNet50, the temporal analysis becomes a critical factor in understanding the motion and detecting potential injuries. This is where LSTM models can be very useful. The sequence of spatial features forms the input for the LSTM model, which learns the temporal dependencies between these features.

However, despite their strong potential, LSTMs are not without drawbacks. Training LSTM models can be computationally intensive and time-consuming, especially when dealing with long sequences of high-dimensional data, such as videos [14]. This might limit the model's utility in scenarios that require real-time analysis and feedback. Addi-tionally, while LSTMs can handle longer sequences than traditional RNNs, they may still struggle with extremely long sequences due to the inherent complexity of learning long-term dependencies.

## 2.3 Stochastic Gradient Descent

Stochastic Gradient Descent (SGD) is an effective and popular optimization algo-rithm that is widely used in machine learning and deep learning tasks[15]. SGD is a var-iant of traditional Gradient Descent (GD) algorithm, characterized by its stochastic nature, wherein instead of using the entire

Journal of Intelligence Technology and Innovation (JITI), 2024, 2(2), 35-52.training dataset to calculate the gradient of the loss function, it uses a single or a subset (mini-batch) of randomly chosen training samples for the same. This drastically reduces the computation cost per iteration and enables the al-gorithm to navigate the parameter space more randomly, thereby assisting in escaping local minima in non-convex problems, such as those common in deep learning.

In the context of sports pose analysis and injury detection, models like the aforemen-tioned LSTM and CNN may face challenges of large computational demand and the risk of overfitting due to a multitude of parameters. SGD serves as a potent tool in addressing these challenges[16]. Firstly, SGD's nature of processing a single or a batch of samples at a time drastically reduces the memory footprint, rendering it feasible for training large-scale deep learning models. Secondly, SGD introduces inherent randomness into the learning process, providing implicit regularization that can alleviate the overfitting problem.

However, SGD also comes with a set of limitations. The foremost issue is its sensitivity to hyperparameters, such as the learning rate and batch size. A small learning rate might make the model slow to converge, while a large one could make the training process unstable. Similarly, the choice of batch size also affects the optimization's quality and speed. Moreover, SGD's performance can be hindered by problems like vanishing/exploding gradients or saddle points[17]. Despite these drawbacks, SGD's benefits, such as computational efficiency and model performance, typically outweigh its limitations, making it a crucial tool in the machine learning toolbox, particularly for large-scale and complex tasks like sports pose analysis and injury detection.

## 3. Materials and Methods

## 3.1 Overview of Our Network

This research investigates the analysis and prediction of athletes' postures and sports injuries through optimizing the ResNet50-BiGRU model using the Sparrow Search Algo-rithm (SSA). The overall process, as depicted in Figure 1, utilizes the ResNet50 model to extract features from each frame of the video. The BiGRU model processes temporal in-formation, while the SSA is used to extract features and reduce dimensions of prepro-cessed data, obtaining keypoint feature information.

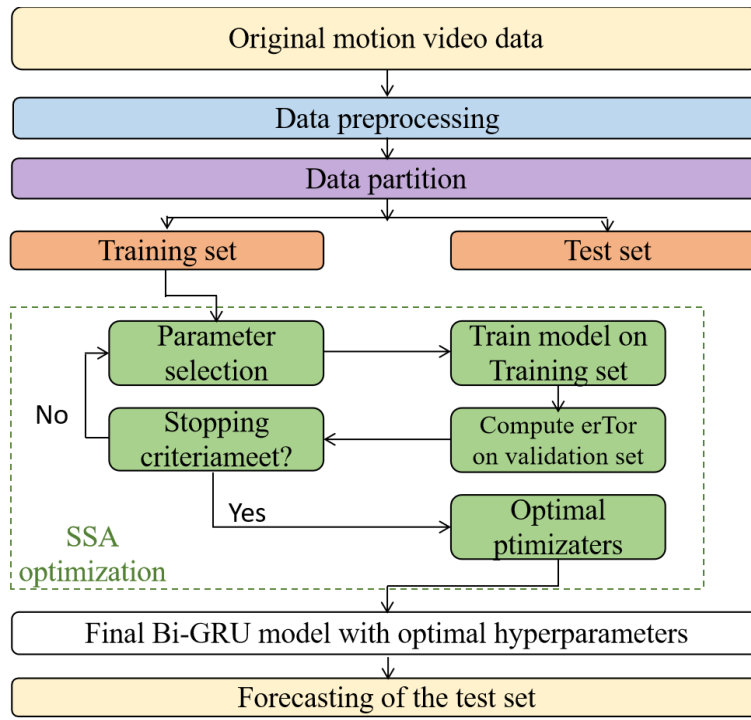

Figure 1. Overall flow chart of the model

The proposed method is based on the ResNet50-BiGRU model, and the specific steps include data preprocessing, feature extraction, model training, model evaluation, and ap-plication. The first step involves preprocessing the athletes' motion video data, followed by grouping and sorting. After the data preprocessing, the SSA is applied to extract fea-tures and reduce dimensions of the processed data, enabling it to better adapt to model training. Subsequently, the ResNet50 and BiGRU models are utilized to extract spatial feature information and temporal sequence information from the data, resulting in a trained model. Finally, the trained model is used with test data for evaluation. It determines whether the athletes' movements in the test data deviate from the norm, aiming to prevent sports injuries.

## 3.2 ResNet50 Model

ResNet50 is a deep residual network notable for its key feature, the residual connection (or residual block)[18]. This form of connection allows for direct connections between inputs and outputs in the network, enhancing the propagation of gradients and effectively mitigating vanishing and exploding gradient problems typically associated with deep networks. An overview of the ResNet50 process can be seen in Figure 2.

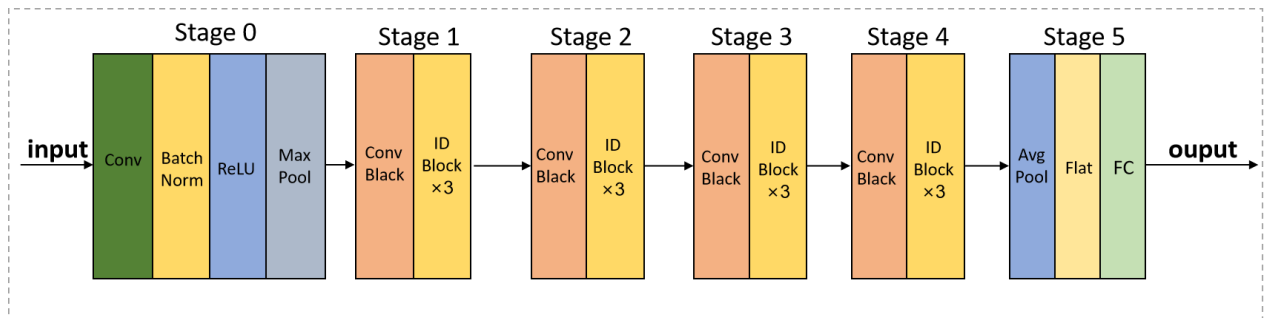

Figure 2. Flow chart of the ResNet50 model

ResNet, short for Residual Network, is a deep neural network architecture, and the variant, ResNet50, denotes the model with approximately 50 layers[19].

The calculation formula[20] of ResNet50 is as follows:

1. Input Layer

This layer accepts the original image data, usually represented as a three-dimensional tensor. If an RGB image is used, the input X will have dimensions [height, width, 3].

2. Convolutional Layer

$$Y_{i,j,k} = \sum_{m,n,l} X_{i+m,j+n,l} \cdot W_{m,n,l,k} + b_k \qquad \text{[Formular 1]}$$

This layer applies a convolution operation on the input, In this equation, $Y_{i,j,k}$ is the output, $X_{i+m,j+n,l}$ is the input, $W_{m,n,l,k}$ is the convolutional filter, and $b_k$ is the bias. The sum is taken over a window of the input, defined by indices $m$, $n$, and $l$.

3. Max Pooling Layer

$$Y_{i,j,k} = \max_{m,n \in P_{i,j}} X_{m,n,k} \qquad \text{[Formular2]}$$

This layer applies a max pooling operation, where the input is divided into rectangular pooling regions, and the maximum value of each region is output. Here, $Y_{i,j,k}$ is the output, $X_{m,n,k}$ is the input, and $P_{i,j}$ defines a pooling region in the input.

4. Residual Block Group

$$Y = F(X, W_i) + X \qquad \text{[Formular 3]}$$

This group contains several residual blocks. Each block contains convolutional layers and uses skip connections. Here, $F(X, W_i)$ represents the weighted layers within the block and $X$ is the input to the block. The function $F$ includes the operations of the convolutional layers in the residual block. The $+X$ represents the skip connection, which adds the original input back to the output of the convolutional layers.

5. Global Average Pooling Layer

$$Y_k = \frac{1}{H \times W} \sum_{i=1}^{H} \sum_{j=1}^{W} X_{i,j,k} \qquad \text{[Formular 4]}$$

This layer applies an average pooling operation to each feature map. Here, $Y_k$ is the output, $X_{i,j,k}$ is the input, $H$ and $W$ are the height and width of the input feature map, respectively.

6. Fully Connected Layer

$$Y = W^T X + b \qquad \text{[Formular 5]}$$

This layer computes the dot product of its input and weights, and adds a bias term. In this equation, $Y$ is the output, $X$ is the input, $X$ are the weights, $b$ is the bias, and $T$ denotes the transpose of a matrix.

## 3.3 BiGRU Model

Bidirectional Gated Recurrent Units (BiGRU) are a type of recurrent neural network (RNN) that are specialized for processing sequences of data[21]. In the context of this study, BiGRU models are utilized to process temporal information obtained from the frames of athlete movement videos, and are particularly useful for identifying and pre-dicting potential athletic injuries.

GRU, or Gated Recurrent Unit, is an RNN architecture[22]. A GRU has two types of gates: a reset gate and an update gate. An overview of the BiGRU process can be seen in Figure 3.

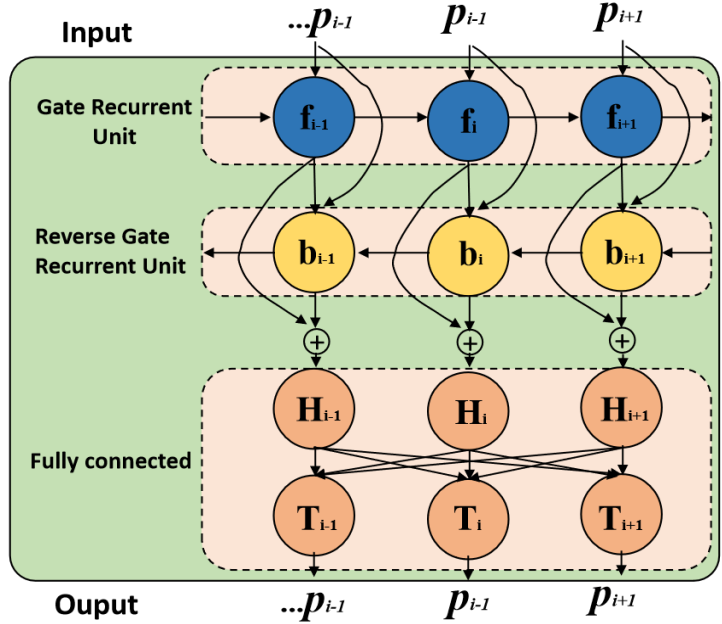

Figure 3. Flow chart of the BiGRU model

The "bidirectional" aspect of BiGRU refers to the way the network processes input data. A BiGRU consists of two GRUs: one processes the data from start to end (forward), while the other processes data from end to start (backward)[23]. The final output for each time step is the concatenation of the forward and backward hidden states.

The workflow of GRU can be divided into four states: update gate, reset gate, candi-date hidden state, and final hidden state, which are summarized by the following equa-tion:

1. Update Gate:
$$z_t = \sigma(W_z \cdot [h_{t-1}, x_t] + b_z) \qquad \text{[Formular 6]}$$

2. Reset Gate:
$$r_t = \sigma(W_r \cdot [h_{t-1}, x_t] + b_r) \qquad \text{[Formular 7]}$$

3. Candidate Hidden State:
$$\tilde{h}_t = \tanh(W \cdot [r_t \odot h_{t-1}, x_t] + b) \qquad \text{[Formular 8]}$$

4. Final Hidden State:
$$h_t = (1 - z_t) \odot h_{t-1} + z_t \odot \tilde{h}_t \qquad \text{[Formular 9]}$$

Where, $z_t$: Update gate at time $t$, determines how much of the past information needs to be passed along to the future. $r_t$: Reset gate at time $t$, determines how much of the past information to forget. $\tilde{h}_t$: Candidate hidden state at time $t$, which contains the candidate information to be transferred to the next state. $h_t$: Final hidden state at time $t$, which contains the information transferred to the next state. $h_{t-1}$: Hidden state at the previous time step. $x_t$: Input at time $t$. $W_z, W_r, W$: Weight matrices for update gate, reset gate and candidate hidden state, learned during training. $b_z, b_r, b$: Bias terms for update gate, reset gate and candidate hidden state, learned during training. $\sigma$: Sigmoid activation function. tanh: Hyperbolic tangent activation function. $\odot$: Denotes element-wise multiplication. [, ]: Denotes concatenation.

In the context of analyzing and predicting athlete movement and potential injuries, BiGRU models are applied to the sequence of feature vectors extracted from each frame by the ResNet50 model. BiGRU models are particularly suitable for this task because of their ability to process temporal sequences and their capacity to capture both past (from the forward GRU) and future (from the backward GRU) context information.

The use of BiGRU allows the model to recognize patterns over time, such as the repetition of a certain movement or the progression of an athlete's posture. This capacity to understand the dynamics of movement over time is crucial in the identification and prediction of potential injury risks.

## 3.4 Sparrow Search Algorithm

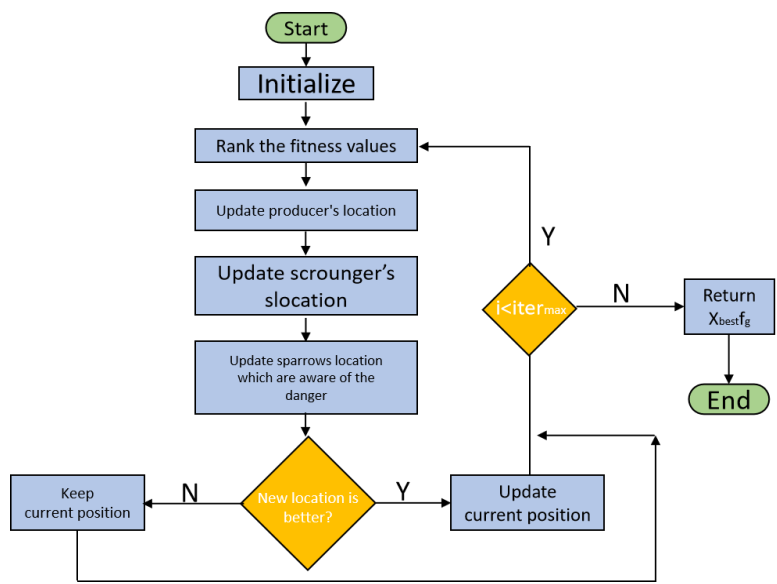

**Figure 4.** Flow chart of the SSA model

Sparrow Search Algorithm (SSA) is a recent, bio-inspired metaheuristic algorithm proposed for optimizing complex problems [24]. It is inspired by the social and foraging behaviors of sparrows in nature. Sparrows engage in various strategies, such as cooperation, competition, and vigilance, to find food while avoiding predators, which is translated into an optimization algorithm to find the best (or optimal) solution to a problem. An overview of the SSA process can be seen in Figure 4.

In the context of the study for posture analysis and sports injury prediction, SSA is used for feature extraction and dimensionality reduction of the preprocessed athlete's motion video data. The algorithm works by iteratively updating potential solutions (positions of "sparrows") to minimize (or maximize) a defined objective function. in this case, a function assessing the quality of the feature set in predicting sports injuries.

In the sparrow search algorithm (SSA) [25], the dynamic behavior of sparrow population is mainly simulated through the following three steps:

**Step1:** Leader position update, and the discoverer are the optimal individuals in the population who search globally in the search space.

$$X_{i,j}^{t+1} = \begin{cases} X_{i,j}^t \cdot \exp(-\frac{i}{\alpha \cdot iter_{max}}) & R_2 < ST \\ X_{i,j}^t + Q \cdot L & R_2 \geq ST \end{cases} \quad \text{[Formular 10]}$$

In the formula, the parameter $t$ denotes the current iteration number, and $iter_{max}$ denotes the maximum number of iterations. $X_{i,j}$ represents the position of the $i$-th sparrow in the $j$-th dimension at the $t$-th iteration. $\alpha$ is a random number within the $(0,1)$ interval. $R_2$ represents the warning value issued by the vigilant sparrow, where $R_2 \in (0,1)$ is the warning safety value and falls within the $(0.5,1)$ interval. If $R_2 < ST$, it implies that no predators have been detected, indicating a safe state of the sparrow population. However, if $R_2 \geq ST$, it signifies that some sparrows have encountered predators. Consequently, all sparrows respond to the danger alarm and urgently relocate to safer zones. In line with equation (10), the discoverer in the subsequent iteration relocates around the current position. The range of variation $Y$ for the movement position value is expressed in equation (11).

$$Y = \exp(\frac{-x}{\alpha \cdot iter_{max}}) \quad \text{[Formular 11]}$$

**Step2:** The position of the Follower is updated, and the follower performs a local search near the solution found by the finder. In each iteration, the follower position is updated as shown in equation (12).

$$X_{i,j}^{t+1} = \begin{cases} Q \cdot \exp(\frac{X_{worst} - X_{i,j}^t}{\alpha \cdot iter_{max}}) & i > n/2 \\ X_{best}^{t+1} + | X_i^t - X_{best}^{t+1} | \cdot A \cdot L & i \leq n/2 \end{cases}, \quad \text{[Formular 12]}$$

**Step3:** All sparrows in the group have the same early warning mechanism, which can be understood as about 10-30% of the sparrows in a group are aware of predators during the foraging process, which triggers the alarm mechanism. The alert's position is updated as shown in equation (13).

$$X_{i,j}^{t+1} = \begin{cases} X_{best}^{t+1} + \beta \cdot | X_{i,j}^t - X_{best}^t | & f_i > f_g \\ X_{best}^{t+1} + k \cdot \frac{1}{f_i - f_w + \varepsilon} & f_i = f_g \end{cases} \quad \text{[Formular 13]}$$

In the aforementioned formula, $k$ represents a random number and $k \in [-1,1]$ acts as a coefficient controlling the random step size. $\beta$ adheres to the Gaussian distribution with a variance of 1 and a mean of 0. $f_i$ denotes the fitness value of the $i$-th sparrow individual. $f_g$ symbolizes the current best fitness value, whereas $f_w$ designates the current worst fitness value. $\varepsilon$ represents the smallest parameter, employed to prevent the denominator from being zero. When $f_i > f_g$, it implies that the individual is on the periphery of the group. If $f_i = f_g$, it indicates that sparrows in the middle of the group have perceived danger, necessitating a scattered flight towards safer locations.

## 4. Results

### 4.1 Experimental Datasets

The research discussed in this paper utilizes the following four datasets to study the athletes' posture and athletic injury data from sports videos:

Sports Pose Datasets [26]: This is a dataset containing human poses in sports scenes. It may contain images of muscular and skeletal changes in various motion scenarios, where joint positions and poses are annotated for each individual. This kind of dataset is very useful for studying the posture changes and motion analysis of the human body in different sports.

Penn Action Datasets [27]: This dataset is used for action recognition and action analysis. It contains action images from YouTube videos, where the action ranges from everyday activities to sports scenes. The images in the dataset all contain labels for categories of muscle changes during exercise.

NTU RGB+D Datasets [28]: This dataset is a multimodal dataset that contains RGB images and depth images. It covers a wide range of human motion and poses, captured in multiple perspectives and environments. This dataset can be used for tasks such as action recognition, behavior analysis, and pose estimation.

Sports-1M Datasets[29]: This is a large-scale sports video dataset that covers a wide range of images of muscular and skeletal changes during exercise. The videos in the dataset contain the movements and poses of the athletes during the competition. It can be used in research such as sports action recognition, action analysis, and video understanding.

### 4.2 Experimental Details

The training process of the ResNet50-BiGRU model optimized with the SSA sparrow search algorithm includes defining the architecture, compiling the model, training the model and saving the model. Each module can be trained independently and combined to form a comprehensive model. This method can effectively improve the accuracy and robustness of the model, so that the model can better cope with the challenges of athletes' posture analysis and sports injury prediction applications.

1. Accuracy:

$$\text{Accuracy} = \frac{\text{TP}+\text{TN}}{\text{TP}+\text{TN}+\text{FP}+\text{FN}} \qquad \text{[Formular 14]}$$

where TP represents the number of true positives, TN represents the number of true negatives, FP represents the number of false positives, and FN represents the number of false negatives.

2. AUC:

AUC, or Area Under the ROC Curve, is a commonly used metric for binary classification problems. The ROC curve plots the True Positive Rate (TPR) against the False Positive Rate (FPR) at various threshold settings.

$$AUC = \int_0^1 ROC(x)dx \qquad \text{[Formular 15]}$$

$ROC(x)$ represents the value of the $ROC$ curve (i.e., the True Positive Rate) at a specific threshold $x$. The integral from 0 to 1 of $ROC(x)dx$ gives the total area under the $ROC$ curve, which is the $AUC$.

## 4.3 Experimental Results and Analysis

In Table 1(a)(b) presents a comparison of different state-of-the-art (SOTA) methods based on various metrics, namely Mean Absolute Error (MAE), Mean Absolute Percentage Error (MAPE), Root Mean Square Error (RMSE), and Mean Square Error (MSE) on four distinct datasets.

Table 1(a). Comparison with different metrics of current SOTA methods

| Model | Datasets | | | | | | | |
|---|---|---|---|---|---|---|---|---|
| | NTU RGB+D datasets | | | | Sports-1M datasets | | | |
| | MAE | MAPE(%) | RMSE | MSE | MAE | MAPE(%) | RMSE | MSE |
| Zadeh et al[30] | 46.05 | 11.27 | 8.21 | 19.61 | 26.37 | 9.14 | 7.84 | 29.18 |
| Huang et al[31] | 46.79 | 14.33 | 4.57 | 24.35 | 49.55 | 9.61 | 4.79 | 15.58 |
| Yang et al[32] | 22.82 | 13.9 | 6.54 | 19.54 | 30.97 | 11.23 | 6.61 | 25.59 |
| Meng et al[33] | 38.68 | 12.83 | 5.01 | 25.94 | 46.01 | 10.43 | 5.45 | 22.72 |
| Oliver et al[34] | 37.37 | 12.11 | 5.25 | 14.52 | 48.92 | 12.04 | 4.31 | 13.05 |
| Henriquez et al[35] | 33.57 | 12.9 | 6.02 | 16.99 | 40.79 | 11.76 | 7.05 | 17.19 |
| Ours | 15.21 | 3.95 | 4.15 | 5.63 | 14.76 | 5.25 | 3.14 | 4.54 |

Table 1(b). Comparison with different metrics of current SOTA methods

| Model | Datasets | | | | | | | |
|---|---|---|---|---|---|---|---|---|
| | SportsPose datasets | | | | Penn Action datasets | | | |
| | MAE | MAPE(%) | RMSE | MSE | MAE | MAPE(%) | RMSE | MSE |
| Zadeh et al | 47.99 | 9.18 | 5.84 | 20.76 | 39.38 | 9.9 | 4.83 | 14.94 |
| Huang et al | 41.29 | 15.48 | 6.81 | 22.8 | 42.67 | 13.56 | 7.51 | 28.79 |
| Yang et al | 25.52 | 9.71 | 5.34 | 18.4 | 24.71 | 14.66 | 8.13 | 17.57 |
| Meng et al | 33.48 | 12.59 | 6.48 | 19.78 | 29.95 | 11.14 | 6.63 | 16.4 |
| Oliver et al | 25.55 | 14.91 | 6.41 | 16.67 | 20.35 | 11.56 | 5.38 | 16.06 |

| Henriquez et al | 40.45 | 14.97 | 7.78 | 28.4 | 47.75 | 12.97 | 6.43 | 13.49 |
| **Ours** | **15.57** | **5.14** | **3.15** | **5.34** | **14.9** | **4.14** | **2.29** | **6.45** |

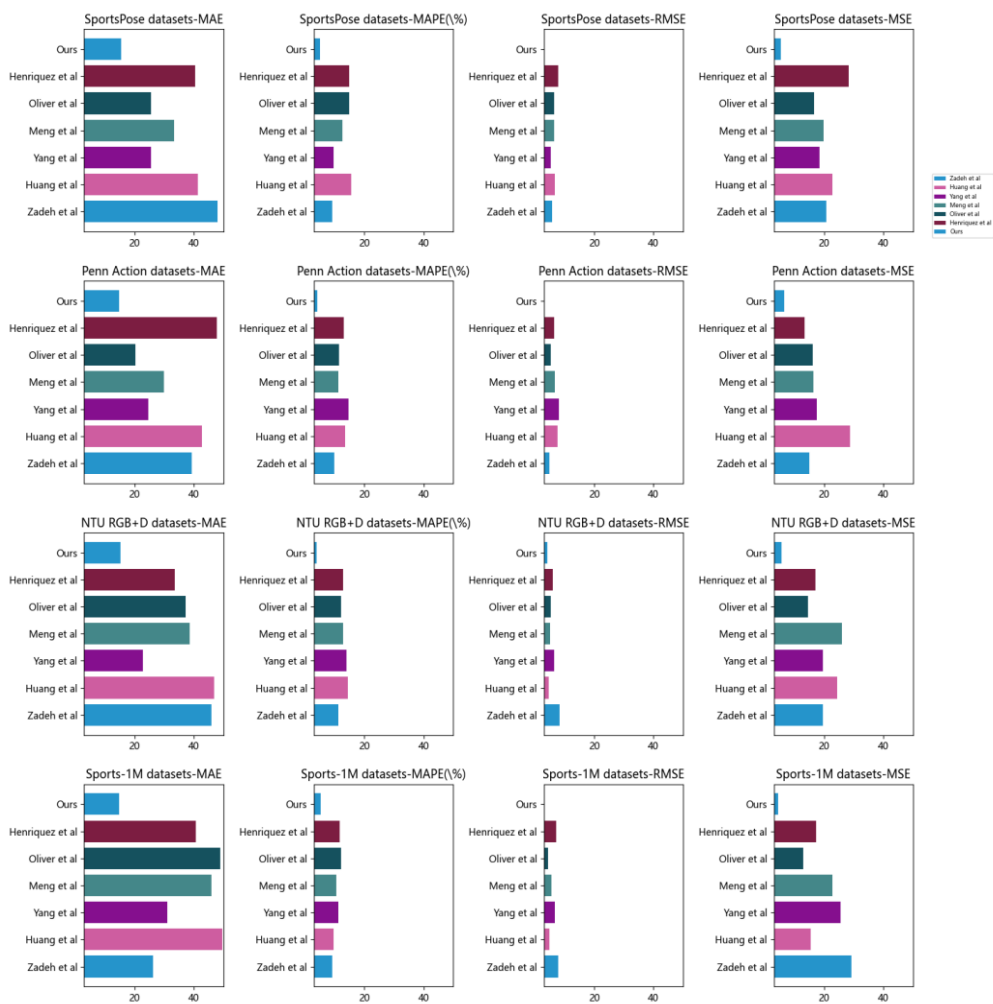

Figure 5. Comparison of different indicators of different models

The key performance indicators of our SSA optimized ResNet50-BiGRU model, including Mean Absolute Error (MAE), Mean Absolute Percentage Error (MAPE), Root Mean Square Error (RMSE), and Mean Square Error (MSE). Compared with other models, our model shows superior performance on these metrics, indicating higher accuracy and robustness of its predictions.

In the figure 5, we compare the seven models of Zadeh et al [30], Huang et al [31], Yang et al [32], Meng et al [33], Oliver et al [34], Henriquez et al [35], our model based on the data of the four error indicators of MAE, MAPE, RMSE and MSE. The data comes from four data sets, among which MAE is one of the important indicators of the prediction model, indicating the average absolute difference between the predicted value and the actual value, and MAPE is the absolute percentage difference between the predicted value and the actual value mean. RMSE is the square root of the mean of the squared differences between predictions and actual observations. MSE is the mean squared of the difference between predicted and actual values. These indicators are the smaller the better, do not have too much error.

Table 2(a). Ablation experiment in CNN module

| Model | Datasets | | | | | | | |
|---|---|---|---|---|---|---|---|---|
| | SportsPose datasets | | | | Penn Action datasets | | | |
| | MAE | MAPE(%) | RMSE | MSE | MAE | MAPE(%) | RMSE | MSE |
| ResNet18 | 20.45 | 9.85 | 6.45 | 26.25 | 25.37 | 13.64 | 4.43 | 29 |
| InceptionV3 | 25.62 | 13.53 | 6.07 | 19.71 | 29.19 | 12.65 | 8.19 | 23.77 |
| EfficientNet | 20.7 | 13.74 | 6.03 | 21.83 | 23.82 | 14.89 | 6.01 | 18.2 |
| ResNet50 | 14.45 | 4.45 | 4.11 | 6.34 | 13.23 | 5.14 | 3.45 | 5.56 |

Table 2(b). Ablation experiment in CNN module

| Model | Datasets | | | | | | | |
|---|---|---|---|---|---|---|---|---|
| | NTU RGB+D datasets | | | | Sports-1M datasets | | | |
| | MAE | MAPE(%) | RMSE | MSE | MAE | MAPE(%) | RMSE | MSE |
| ResNet18 | 48.29 | 9.75 | 6.36 | 22.75 | 41.03 | 11.54 | 4.9 | 27.01 |
| InceptionV3 | 31.2 | 13.71 | 8.17 | 19.67 | 46 | 10.87 | 6.86 | 20.75 |
| EfficientNet | 42.74 | 13.68 | 2.37 | 13.4 | 50.06 | 9.68 | 8.01 | 29.67 |
| ResNet50 | 16.45 | 4.31 | 4.15 | 5.61 | 15.24 | 4.67 | 3.56 | 5.13 |

Table 2(a)(b) presents study results for four models—ResNet18, InceptionV3, EfficientNet, and our proposed mode. In the experiment, our model consistently outperforms the others, showing lower MAE, MAPE, RMSE, and MSE across all datasets, signifying more accurate and reliable predictions. For instance, on the SportsPose dataset, our model's MAE of 14.45 outshines the 20.45, 25.62, and 20.7 of ResNet18, InceptionV3, and EfficientNet, respectively. Figure 6 shows the visualization results of Table 2(a)(b). It can be clearly seen from the visualization that the ResNet50 module in our model has smaller errors than other modules.

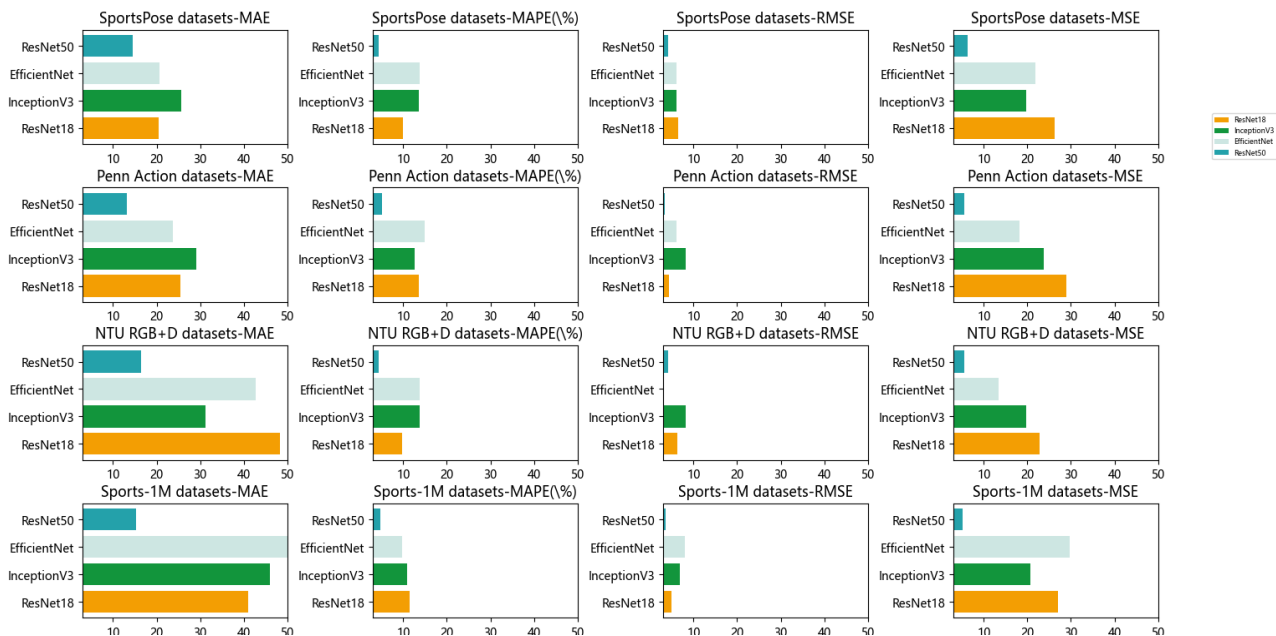

Figure 6. Comparison of different indicators of different models

These results affirm the superior generalization ability of our model across diverse datasets, offering more precise and consistent predictions. The model's high performance is attributed to its effective learning mechanisms, enabling better pattern capture and representation. Thus, the experiment underscores the effectiveness and superiority of our model in these tasks.

Table 3. Ablation experiment in CNN module

| Model | Datasets | | | | | | | |
| --- | --- | --- | --- | --- | --- | --- | --- | --- |
| | SportsPose datasets | | Penn Action datasets | | NTU RGB+D datasets | | Sports-1M datasets | |
| | Accuracy(%) | AUC(%) | Accuracy(%) | AUC(%) | Accuracy(%) | AUC(%) | Accuracy(%) | AUC(%) |
| LSTM | 92.65 | 87.04 | 97.14 | 81.11 | 96.62 | 82.63 | 85.7 | 98.14 |
| CNN | 93.79 | 93.06 | 80.45 | 83.16 | 89.99 | 97.13 | 85.53 | 89.75 |
| Transformer | 92.57 | 90.87 | 90.9 | 83.37 | 98.45 | 97.79 | 85.35 | 92.02 |
| BiGRU | 96.44 | 95.56 | 95.45 | 94.34 | 95.33 | 93.55 | 97.45 | 95.76 |

Table 3 summarizes the ablation study results, tested against LSTM, CNN, Transformer, and BiGRU models on four diverse datasets: SportsPose, Penn Action, NTU RGB+D, and Sports-1M. Figure 7 shows the visualization results of Table 3. It can be clearly seen from the visualization that the BiGRU module in our model has the best AUC(%) value on the SportsPose dataset and the Penn Action dataset, and ranks well in the NTU RGB+D dataset and Sports-1M dataset.

Our proposed model, demonstrates the highest performance across all datasets in terms of both accuracy and AUC. For example, on the SportsPose dataset, BiGRU achieved an accuracy of 96.44% and AUC of 95.56%, outperforming LSTM (92.65%, 87.04%), CNN (93.79%, 93.06%), and Transformer (92.57%, 90.87%). Similar patterns are seen in the other datasets, reinforcing the superiority of BiGRU. This superior performance can be attributed to the bidirectional structure of BiGRU, which allows it to learn complex temporal dependencies from both past and future context, making it more effective for tasks like motion pose estimation and injury prediction. Therefore, these results underscore BiGRU's effectiveness and potential in tasks relating to athlete performance and injury prevention.

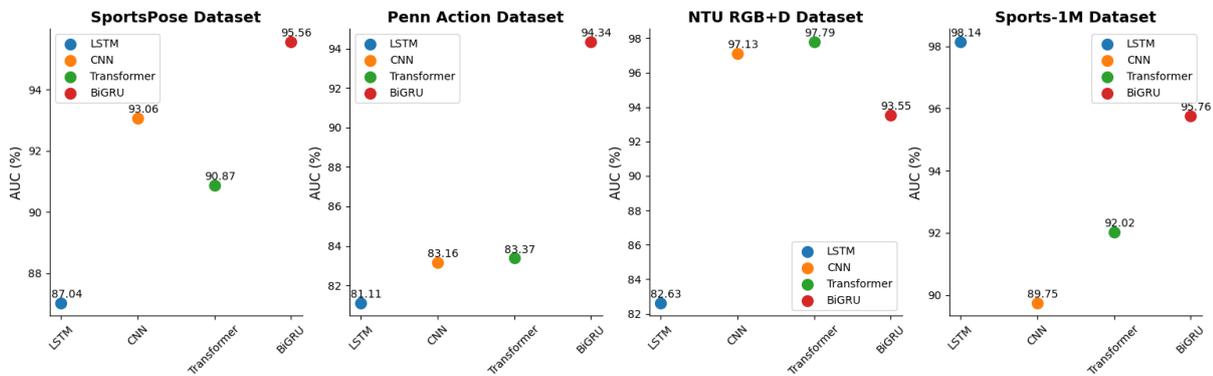

Figure 7. Comparison of different indicators of different models

## 5. Conclusions

In conclusion, our research has made a substantial contribution to the domain of anomaly detection in images, particularly in monitoring the intricate musculoskeletal changes during physical exercise. Our ResNet50-BiGRU model, optimized with the Sparrow Swarm Algorithm, emerges as a robust and accurate tool for predicting athletes' performance and identifying potential exercise-related injuries. This holds profound practical implications for injury prevention strategies and enhancing sports injury management. By providing precise forecasts of potential injuries, our model enables the creation of personalized training regimens and preventive measures that account for the unique muscular and skeletal adaptations during exercise. This not only mitigates the risk of injuries but also optimizes athletic performance.

As we look forward, our research lays the groundwork for further advancements, encouraging the exploration of error-resilient labeling techniques and algorithmic enhancements to elevate the efficacy and interpretability of our model. Future research will explore further optimization techniques to improve model efficiency and scalability, enabling real-time analysis and feedback. Additionally, we plan to extend our approach to other domains requiring precise anomaly detection and predictive analysis, such as medical diagnostics and industrial monitoring. The journey towards more accurate and reliable image anomaly detection continues, offering a brighter and safer future for athletes and individuals engaged in physical activities.

## Author contributions

## Data availability

The data that support the findings of this study are available on request from the corresponding author. The data are not publicly available due to privacy or ethical restrictions.

## Declaration of competing interest

The authors declare that they have no known competing financial interests or personal relationships that could have appeared to influence the work reported in this paper.